\documentclass[runningheads]{llncs}
\usepackage[T1]{fontenc}
\usepackage{graphicx}
\usepackage[misc]{ifsym}
\usepackage{orcidlink}
\usepackage{booktabs}
\usepackage{multirow}
\usepackage{enumitem}
\usepackage{siunitx}
\usepackage{url}

\begin{document}
\title{Does the Prompt-based Large Language Model Recognize Students' Demographics and Introduce Bias in Essay Scoring?}
\titlerunning{Large Language Model Bias in Essay Scoring}
%
\author{Kaixun Yang\orcidlink{0009-0005-0308-0258} \and
Mladen Rakovi\'{c}\orcidlink{0000-0002-1413-1103} \and
Dragan Ga\v{s}evi\'{c}\orcidlink{0000-0001-9265-1908} \and
Guanliang Chen\textsuperscript{(\Letter)}\orcidlink{0000-0002-8236-31335}
}
\authorrunning{K. Yang et al.}
%
\institute{Centre for Learning Analytics, Monash University, Melbourne, Australia \\ \email{\{Kaixun.Yang1, Mladen.Rakovic, Dragan.Gasevic, Guanliang.Chen\}@monash.edu}}
\maketitle              
\begin{abstract}
Large Language Models (LLMs) are widely used in Automated Essay Scoring (AES) due to their ability to capture semantic meaning. Traditional fine-tuning approaches required technical expertise, limiting accessibility for educators with limited technical backgrounds. However, prompt-based tools like ChatGPT have made AES more accessible, enabling educators to obtain machine-generated scores using natural-language prompts (i.e., the prompt-based paradigm). Despite advancements, prior studies have shown bias in fine-tuned LLMs, particularly against disadvantaged groups. It remains unclear whether such biases persist or are amplified in the prompt-based paradigm with cutting-edge tools. Since such biases are believed to stem from the demographic information embedded in pre-trained models (i.e., the ability of LLMs' text embeddings to predict demographic attributes), this study explores the relationship between the model’s predictive power of students' demographic attributes based on their written works and its predictive bias in the scoring task in the prompt-based paradigm. Using a publicly available dataset of over 25,000 students' argumentative essays, we designed prompts to elicit demographic inferences (i.e., gender, first-language background) from GPT-4o and assessed fairness in automated scoring. Then we conducted the multivariate regression analysis to explore the impact of the model's ability to predict demographics on its scoring outcomes. Our findings revealed that (i) Prompt-based LLM can somewhat infer students’ demographics, particularly their first-language backgrounds, from their essays; (ii) Scoring biases are more pronounced when the LLM correctly predicts students’ first-language background than when it does not; (iii) Scoring error for non-native English speakers increases when the LLM correctly identifies them as non-native.

\end{abstract}
\section{Introduction} \label{sec:intro}
In recent years, Large Language Models (LLMs) have undergone rapid advancements. These technologies have demonstrated significant potential for application in various educational contexts, such as assistance for writing \cite{lee2022coauthor}, generation of learning materials \cite{binhammad2024investigating}, and virtual tutoring \cite{garcia2024review}. Among these educational tasks, writing assessment is particularly important, as effective evaluation of student writing plays a crucial role in improving writing performance and helping students develop their writing skills \cite{deane2022importance}. Due to the time-consuming and labor-intensive nature of manual writing assessment, researchers are exploring Automated Essay Scoring (AES), a process that employs artificial intelligence techniques to assign scores to student-written essays \cite{burstein2004automated}. Pre-trained LLMs (e.g., BERT) have commonly been used for AES by following the fine-tuning paradigm \cite{rodriguez2019language,yang2020enhancing}, which involves using a small set of labeled essays to adapt the LLMs for the specific scoring task at hand. While these fine-tuned models have demonstrated effectiveness in AES, their development typically requires educational practitioners to possess expertise in programming and machine learning. Additionally, training these models demands considerable time and computational resources, posing a significant challenge for many educators \cite{mayfield2020should}. The emergence of prompt-based LLMs, such as ChatGPT, has enabled non-tech savvy educators to access advanced AI technologies to support their teaching practices \cite{yan2024promises}. These conversational LLMs allow users to provide natural-language instructions to guide the models in completing specific tasks (i.e., via prompts). Recent research has investigated AES using LLMs through prompt-based methods, comparing their accuracy with fine-tuning approaches. While prompt-based methods exhibit slightly lower accuracy, they still achieve satisfactory performance when utilizing prompting techniques such as few-shot learning \cite{xiao2025human,stahl2024exploring}.

Since LLMs are typically trained on large corpus that may contain historical biases and discriminatory content towards minority groups, the outputs they generate can perpetuate social biases, potentially harming marginalized communities \cite{ganguli2022red}. A recent study has demonstrated that stereotypical associations are encoded in BERT when performing pronoun resolution tasks \cite{de2021stereotype}. Moreover, such biases have been shown to negatively impact the predictive fairness of downstream educational text classification tasks \cite{sha2022bigger}, where models exhibit favoritism toward certain student groups based on inherent or acquired characteristics (e.g., gender and first-language background) \cite{mehrabi2021survey}. Several studies have evaluated the predictive bias of AES when utilizing LLMs \cite{yang2024unveiling,schaller2024fairness,kwako2024can}; however, these studies primarily focus on fine-tuning paradigms. Given that prompt-based tools like ChatGPT are built on transformer architectures similar to conventional LLMs such as BERT, and considering the unique nature of the prompt-based paradigm, where models rely on natural-language instructions and a few in-context learning examples, it is possible that such biases persist or even become exacerbated due to the model’s limited "training" in performing specific tasks properly. 

To bridge the gap in understanding whether recent prompt-based LLMs continue to exhibit biases in AES tasks, we build on prior research that explores the connection between bias in fine-tuned LLMs and subsequent predictive fairness in downstream tasks \cite{kwako2024can,sha2022bigger}. Specifically, we set out to explicitly measure the ability of LLMs to infer students’ demographic attributes through the linguistic features of their written essays. Furthermore, we aimed to analyze the relationship between LLMs' ability to predict demographic attributes and its predictive bias in AES tasks. Our study can guide strategies for mitigating biases in prompt-based LLMs, promoting fairer scoring for students from diverse backgrounds and advancing educational equity. Formally, our research addresses the following \textbf{R}esearch \textbf{Q}uestions:

\begin{enumerate}[label=\bfseries RQ\arabic*,leftmargin = 30pt]
    \item To what extent can an LLM discern students' demographic attributes based on the linguistic characteristics in their written essays?
    \item To what extent does the LLM's ability to predict students' demographic attributes relate to their bias in scoring students' essays?
\end{enumerate}

To answer the RQs, we chose a publicly available dataset comprising over 25,000 argumentative essays, each assigned a holistic score, spanning 15 distinct topics. This dataset also includes various demographic attributes of students, such as gender, first-language background, and race. We employed the state-of-the-art LLM model, GPT-4o, which has been widely used in recent AES research \cite{xiao2025human,stahl2024exploring}.To answer RQ1, we prompted GPT-4o to infer students' gender and first-language background based on their essays. This allowed us to assess the model's ability to infer demographic attributes using linguistic features present in students' writing. To answer RQ2, we prompted GPT-4o to score the essays and evaluated its predictive fairness using multiple evaluation metrics. Specifically, we analyzed how the model's ability to predict students' demographics (identified in RQ1) affected scoring outcomes. To further quantify these relationships, we conducted multivariate regression analysis. All the prompts were designed by including the effective prompting techniques used in recent literature (e.g., few-shot learning, chain-of-thought reasoning, and role-assigned prompting), as detailed in Section \ref{sec:method}. Through extensive analysis, this study contributed with the following key findings: (i) Prompt-based LLMs can infer students' demographics, particularly their first-language backgrounds, from their essays. 
When GPT-4o delivered explicit prediction labels of students' demographics (i.e., excluding those receiving the label of 'uncertain'),
it achieved an accuracy of approximately 0.86–0.96 for gender and 0.75–0.87 for first-language background; (ii) Scoring biases persist regardless of whether prompt-based LLMs accurately identify students' language backgrounds. However, these biases are more pronounced when the LLM correctly predict students' first-language backgrounds; (iii) Statistically significant interaction terms with positive coefficients in the regression models for particular essay sets suggest that the scoring error for non-native speakers increases when the model correctly identifies them as non-native.

\section{Related Work} \label{sec:background}
\subsection{Automated Essay Scoring}
Over the past decades, AES has undergone rapid development. According to the existing literature \cite{zesch2015task,chen2013automated,dong2017attention,taghipour2016neural,yang2020enhancing,yang2024unveiling}, these approaches can generally be grouped into four categories: traditional machine learning-based methods, deep learning-based methods, methods based on fine-tuning LLMs, and methods based on prompting LLMs.

\noindent \textbf{Traditional machine learning-based methods.} To enable an AES model to accurately assess essay quality, these methods typically placed considerable emphasis on manually designing meaningful textual features to serve as input for training traditional machine learning models (e.g., Bayesian Linear Regression, Random Forests, and Support Vector Machines (SVM)). For instance, Zesch et al. \cite{zesch2015task} improved the training of an AES model using SVM by leveraging a comprehensive feature set that included key linguistic attributes essential for evaluating essay quality, such as word n-grams, cohesion features, and syntactical features. Similarly, Chen et al. \cite{chen2013automated} employed a variety of linguistic and statistical features (e.g., lexical features, grammar and fluency features, and syntactical features) to train a rank-based SVM rating model, where the agreement between human raters and the model is explicitly integrated into the loss function.

\noindent \textbf{Deep learning-based methods.}
Manually designing features is a laborious process, leading some researchers to explore the use of deep learning models to automate the extraction of features from raw textual data. For example, Taghipour et al. \cite{taghipour2016neural} utilized Convolutional Neural Networks to identify local textual dependencies and Long Short-Term Memory networks to capture sequential dependencies, leveraging these extracted features to score essays. Similarly, Dong et al. \cite{dong2017attention} employed hierarchical network structures to capture dependencies at the word and sentence levels, incorporating attention mechanisms into Recurrent Neural Networks to highlight key words or sentences to score essays.

\noindent \textbf{Methods based on fine-tuning LLMs.} With the emergence of pre-trained LLMs (e.g., BERT), which are trained on large text corpora to learn general linguistic features, some researchers have begun leveraging these advancements to score essays more accurately by further fine-tuning these models on a smaller set of labeled essays to adapt these models to the specific scoring task. For example, Rodriguez et al. \cite{rodriguez2019language} fine-tuned BERT to generate essay embeddings for subsequent scoring. Similarly, Yang et al. \cite{yang2020enhancing} introduced a hybrid loss function that integrates dynamically weighted mean square error loss with batch-wise ListNet loss, enhancing scoring accuracy during fine-tuning BERT.

\noindent \textbf{Methods based on prompting LLMs.} Fine-tuning LLMs demands substantial computational resources and often yields only holistic scores, which oftentimes fails to provide detailed explanations. To address this, recent research has investigated prompt-based LLMs (e.g., ChatGPT) for essay scoring and explanation generation. These approaches typically rely on natural-language prompts to guide LLMs in performing the scoring task. For instance, Xiao et al. \cite{xiao2025human} proposed various prompting strategies, such as including or excluding detailed rubric contexts and utilizing zero-shot or few-shot in-context learning strategies, to guide LLMs in scoring student essays and providing feedback. Similarly, Stahl et al. \cite{stahl2024exploring} examined the effectiveness of various task instruction types and few-shot learning during the prompt design phase for essay scoring.

\subsection{Bias in Automated Essay Scoring}
The importance of algorithmic bias in AES was recognized as early as 2012 \cite{williamson2012framework}. However, it was not until recent years that several studies were conducted to explore the biases in existing AES algorithms. For example, Litman et al. \cite{litman2021fairness} analyzed the biases of three AES models, highlighting that different models exhibit distinct biases related to students' gender, race, and socioeconomic status. Yang et al. \cite{yang2024unveiling} examined nine widely used AES methods, assessing their performance across seven metrics on an open-source dataset. Their findings revealed that topic-specific models tend to display greater bias towards students of varying economic backgrounds compared to cross-topic models. Schaller et al. \cite{schaller2024fairness} investigated shallow learning, deep learning, and LLMs using both balanced and skewed training data subsets. They found that models trained on skewed data from students with higher or lower cognitive abilities showed no bias but suffered from significantly reduced accuracy for students outside the training set. However, these studies have primarily focused on the fine-tuning paradigm in using LLMs. To the best of our knowledge, only one study has explored AES fairness in prompt-based LLMs. Yancey et al. \cite{yancey2023rating} assessed GPT-4’s scoring performance across different genders and L1 language groups, concluding that bias did not significantly vary based on gender or L1. Across all the studies mentioned above, while they demonstrated the presence of predictive bias, they did not explicitly investigate to what extent the observed scoring bias is related to LLMs' ability to predict students' demographic attributes, while an LLM's ability to predict students' demographic attributes has been recognized to be somewhat related to its predictive bias towards disadvantaged users \cite{sha2022bigger}. As a result, these studies provide limited insights for developing effective debiasing strategies to address predictive bias issues in existing AES systems. To our knowledge, only one study \cite{kwako2024can} has attempted to investigate such relationships in AES. The authors fine-tuned XLNet to score essays written by students from different demographic backgrounds and then examined whether XLNet’s hidden states could predict students’ demographic attributes. Their findings provided evidence that demographic group differences were embedded in the model’s hidden layers. However, their study focused on fine-tuning LLMs. Whether and how such relationships exist in prompt-based LLMs (especially the latest ones like GPT-4o) remains  unexplored. Our study fills this gap by prompting LLMs to infer students’ demographic attributes from their essays, score the essays, and quantify the relationship between LLMs' ability to predict these attributes and biases in essay scoring.

\section{Method} \label{sec:method}
\subsection{Dataset}
The PERSUADE 2.0 corpus comprises over 25,000 argumentative essays written by U.S. students in grades 6 through 12 across 15 different topics \cite{crossley2024large}. The dataset includes holistic essay scores assigned by trained human raters based on a standardized scoring rubric used in the Scholastic Aptitude Test (SAT). These holistic scores range from 1 to 6, with higher scores indicating better essay quality. Additionally, the dataset contains various demographic attributes of the students, such as gender, race, and first-language background. The PERSUADE 2.0 corpus  has been utilized in several studies on AES fairness \cite{yang2024unveiling,kwako2024can}, further demonstrating its usability and quality. However, since some writing topics are source-based writing and the dataset does not include the source articles, and some topics lack demographic information, we retained only 6 independent writing topics for our experiments. The dataset was publicly released in a Kaggle competition and was already split into a 40:60 ratio for training and testing with stratified random sampling. As a result, we conducted our experiments and evaluation on 60\% of the data, while the remaining 40\% was used as a pool for selecting few-shot examples, as detailed in Section \ref{sec:pd}. The distribution of the 60\% experimental dataset is presented in Table \ref{table1}. Given the dataset's imbalance in first-language background distribution, our study applied techniques to mitigate its impact, as detailed in Sections \ref{sec:ed} and \ref{sec:em}. Previous research has shown that students' essays display unique linguistic features based on gender and first-language background \cite{yoon2021interactions,jones2007discourses}, which motivated us to investigate whether LLMs were capable of discerning these demographic attributes solely by analyzing student written essay. 

\begin{table*}[hbt!]
\begin{center}
\caption{Distribution of student demographics, with 'Native/Non-Native' indicating whether students are Native or Non-Native English speakers.}
\label{table1}
\resizebox{0.9\textwidth}{!}{
\begin{tabular}{@{}c|c|rr|rr@{}}
\toprule
\multirow{2}{*}{Essay Set} & \multirow{2}{*}{Topic} & \multicolumn{2}{c|}{Language Background} & \multicolumn{2}{c}{Gender} \\ \cmidrule(l){3-6} 
 &  & Native & Non-Native & Male & Female \\ \midrule
1 (n=875) & Summer projects & 849 & 26 & 442 & 433 \\
2 (n=839) & Mandatory extracurricular activities & 779 & 60 & 412 & 427 \\
3 (n=813) & Cell phones at school & 781 & 32 & 372 & 441 \\
4 (n=807) & Grades for extracurricular activities & 770 & 37 & 375 & 432 \\
5 (n=760) & Community service & 728 & 32 & 353 & 407 \\
6 (n=1,498) & Distance learning & 1,124 & 374 & 743 & 755 \\ \bottomrule
\end{tabular}
}
\end{center}
\end{table*}

\subsection{Prompting Design} \label{sec:pd}
To ensure the independence of experimental results, we created three independent prompts and submitted them to GPT-4o in separate sessions: one for inferring gender, one for inferring language background, and one for scoring. To enhance the quality of LLMs responses, we reviewed literature on effective prompting techniques \cite{chen2023unleashing,sahoo2024systematic} as well as previous studies on using LLMs for AES \cite{xiao2025human,stahl2024exploring}. The prompting techniques we employed include:

\smallskip
\noindent \textbf{Role-assigned Prompting.} Explicitly assigning a role or persona to LLMs (e.g., educator) enhances contextual understanding and enables responses that align with specific expectations \cite{olea2024evaluating}. For instance, a prompt such as \textit{"You are an educator with expertise in grading essays for students in Grades 6 through 12 in the U.S."} helps tailor LLM’s output to the desired context.

\smallskip
\noindent \textbf{Few-Shot Prompting.} Providing LLMs with a few input-output examples helps them understand a given task. Even a small number of high-quality examples has been shown to enhance LLMs' performance on complex tasks compared to no demonstrations \cite{sahoo2024systematic}. Prior studies \cite{yancey2023rating,xiao2025human} in AES tasks indicate that increasing the number of examples does not always lead to better results, often exhibiting diminishing marginal improvement. Therefore, following previous research, we selected 3 samples for this study. We selected few-shot samples by randomly choosing essays from the 40\% portion of dataset mentioned above, ensuring one high-scoring (i.e., score of 6), one medium-scoring (i.e., score of 3), and one low-scoring essay (i.e., score of 1). This selection method has been proven effective in previous AES studies \cite{yoshida2024impact}.

\smallskip
\noindent \textbf{Chain of Thought (CoT) Prompting.} 
Prompting LLMs in a manner that encourages coherent, step-by-step reasoning processes can effectively generate more structured and thoughtful responses compared to traditional prompts \cite{wang2022self}. Following previous studies \cite{xiao2025human,stahl2024exploring,li2023can}, we structured the prompts by breaking tasks into a step-by-step format to implement the CoT approach.

\smallskip
\noindent \textbf{Triple Quotes.} Using triple quotes to separate different parts of a prompt or encapsulate multi-line strings (e.g., the essay text) can enhance the model's understanding of instructions \cite{chen2023unleashing}.

\smallskip
\noindent \textbf{Output Formatting.} Designing prompts that direct LLMs to produce responses in a structured, organized format can ensure models produce all the required output as we expect \cite{chen2024systematic}. In our study, we instructed LLMs to generate responses in JSON format, including both predictions and explanations.

Particularly, for the prompt designed for the scoring task, we also integrated the scoring rubrics into the prompt, as including rubrics have demonstrated improved performance in previous studies \cite{xiao2025human}. All the designed prompts are available in the digital appendix \footnote{\url{https://bit.ly/AIED25_Appendix}}.

\subsection{Experimental Design} \label{sec:ed}
Our study is primarily divided into two phases.  

\noindent\textbf{Phase I}: We requested LLM to infer students' gender and first-language background based on their essays. For gender, we instructed LLM to select from ('Male', 'Female', 'Uncertain'), and for language background, it selected from ('Native', 'Non-native', 'Uncertain'). The coverage rate refers to the proportion of cases where LLM provides a definitive response (i.e., not 'Uncertain'). We classified all correct predictions as \texttt{'Correct'}, while incorrect predictions and cases marked as 'Uncertain' are categorized as \texttt{'Unreliable'}.

\noindent\textbf{Phase II}: We then requested LLM to score a student's essay in a separate session. We calculated bias measures towards different demographic student cohorts in each of the 'Correct' and 'Unreliable' groups to compare whether the bias differs between these two groups. Notably, bias was evaluated based on gender and first-language background, respectively, depending on whether the LLM's prediction pertains to gender or first-language background. To further quantify this relationship, we performed the weighted multivariate regression analysis using inverse probability weighting, which assigns higher weights to minority attribute samples to address imbalance. The dependent variable was the error in the LLM's predicted essay scores, while the independent variables included demographics (e.g., gender and first-language background), correctness status (\texttt{Correct} vs. \texttt{Unreliable}), and interaction terms between demographics and correctness status. 

\subsection{Evaluation Metrics} \label{sec:em}
To assess the accuracy of LLM in inferring students' demographics, we utilized \textbf{accuracy} and \textbf{weighted F1-scores}. Specially, the weighted F1-score adjusts for class imbalance by assigning weights to each class's F1-score, making it particularly suitable for our highly imbalanced dataset, as shown in Table \ref{table1}. To assess the accuracy of LLM in scoring student essays, we used \textbf{Quadratic Weighted Kappa (QWK)}, a widely used metric in AES research that measures agreement between two raters while accounting for the extent of disagreement \cite{lagakis2021automated}. To assess the bias of LLM in scoring student essays, we built upon previous studies and adopted three key metrics to evaluate the extent to which an AES model's predictive errors can be attributed to students' demographic traits \cite{yang2024unveiling,loukina2019many,litman2021fairness}. \textbf{Overall Score Accuracy (OSA)} measures bias by analyzing how much of the variance between the AES model’s predicted scores and the actual scores can be explained by students’ demographic attributes. \textbf{Overall Score Difference (OSD)} specifically identifies whether the AES model tends to overestimate or underestimate scores for certain student groups. \textbf{Conditional Score Difference (CSD)} further accounts for students' language proficiency, which is approximated using their ground-truth essay scores. Beyond these three metrics, we also assessed fairness from a scale perspective using \textbf{Mean Absolute Error Difference (MAED)} \cite{yang2024unveiling}, which quantifies the disparity by comparing the \textbf{Mean Absolute Error (MAE)} between different student groups. For gender, the reference group consists of female students, while for first-language background, the reference group includes non-native English speakers. Since the calculation of OSA, OSD, and CSD requires constructing regression models, we applied inverse probability weighting to mitigate the effects of class imbalance in our selected dataset.

\subsection{Implementation}
We selected the state-of-the-art LLM model, GPT-4o, which has demonstrated superior performance in previous AES studies \cite{xiao2025human,stahl2024exploring,yancey2023rating}. All experiments were conducted using the GPT-4o API. To ensure reproducibility, we set the \textit{Temperature} parameter to 0. Each experiment was run five times with paraphrased prompts by other prompt-based LLMs (i.e., Gemini and Claude) to ensure generalizability. For demographic predictions, we applied majority voting, while essay scoring was determined by averaging the results. The experiments are conducted independently for each of the six essay sets to ensure the validity of the results.

\section{Results} \label{sec:result}

\subsection{Results on RQ1}
The demographic prediction results are presented in Table \ref{table2}. Firstly, LLM demonstrated a high coverage rate for first-language background (approximately 97\%–99\%) but a much lower coverage rate for gender (around 4\%–13\%). This indicates that LLM was more effective at recognizing linguistic patterns associated with different language backgrounds than those related to gender differences. Secondly, for all covered samples, LLM demonstrated high performance in accurately classifying both first-language background and gender. For gender, despite the low coverage rate, the model achieved an accuracy of approximately 0.86–0.96 and a weighted F1-Score of around 0.91–0.96. This suggests that LLM required highly distinct linguistic patterns written by students of different genders to make accurate predictions. Similarly, for first-language background, LLM performed well, achieving an accuracy of about 0.79–0.86 and a weighted F1-Score ranging from 0.86 to 0.92.

\begin{table*}[hbt!]
\begin{center}
\caption{Demographics prediction results: The Coverage represents the percentage of predictions that are not classified as 'uncertain.'}
\label{table2}
\resizebox{0.8\textwidth}{!}{
\begin{tabular}{@{}c|ccc|ccc@{}}
\toprule
\multirow{2}{*}{\begin{tabular}[c]{@{}c@{}}Essay \\ Set\end{tabular}} & \multicolumn{3}{c|}{Language Background} & \multicolumn{3}{c}{Gender} \\ \cmidrule(l){2-7} 
 & Coverage & Weighted F1 & Accuracy & Coverage & Weighted F1 & Accuracy \\ \midrule
1 & 0.998 & 0.921 & 0.868 & 0.064 & 0.947 & 0.947 \\
2 & 0.997 & 0.881 & 0.791 & 0.135 & 0.964 & 0.961 \\
3 & 0.985 & 0.912 & 0.794 & 0.044 & 0.941 & 0.857 \\
4 & 0.970 & 0.910 & 0.753 & 0.095 & 0.908 & 0.900 \\
5 & 0.976 & 0.883 & 0.841 & 0.062 & 0.936 & 0.931 \\
6 & 0.998 & 0.866 & 0.862 & 0.078 & 0.949 & 0.924 \\ \midrule
Average & 0.987 & 0.896 & 0.818 & 0.080 & 0.941 & 0.920 \\ \bottomrule
\end{tabular}
}
\end{center}
\end{table*}

\subsection{Results on RQ2}
Table \ref{table3} presents the accuracy and bias results across various essay sets. In addition to reporting the QWK for each essay set, we also calculated QWK for the 'Correct' and 'Unreliable' groups based on gender and first-language background, respectively. Overall, the QWK is 0.629, with scores ranging from 0.456 (Essay Set 1) to 0.606 (Essay Set 2). According to the QWK standard \cite{doewes2023evaluating}, this indicates a moderate level of agreement between the LLM and human raters. Furthermore, the QWK score range is consistent with previous studies on AES using prompt-based LLMs \cite{xiao2025human,stahl2024exploring}. Interestingly, the QWK for the 'Correct' groups based on first-language background is consistently higher than that of the 'Unreliable' groups across all essay sets. This suggests that the LLM's scoring accuracy is higher when it successfully predicts a student's first-language background compared to when it fails to do so. In contrast, the QWK for the 'Correct' and 'Unreliable' groups based on gender remains similar.

Regarding language background, both the "Unreliable" and "Correct" groups exhibit multiple instances of identified unfairness (i.e., the absence of 'ns' labels in certain cells) across all essay sets. However, when analyzing the dataset as a whole, an interesting pattern emerges: scoring bias is more pronounced when the LLM correctly predicts students' first-language background than when it does not. This is reflected in larger absolute values of MAED and statistically significant results for OSA, OSD, and CSD. Notably, in the "Correct" groups, MAED is negative, indicating that when the LLM correctly predicts students' first-language background, non-native English writers experience greater scoring errors compared to native writers. In contrast, when the LLM fails to predict students' first-language background correctly, non-native English writers experience fewer scoring errors. For gender, fairness metrics showed minimal disparities, with most cells being non-significant. Even when significant, the values remained close to zero, indicating that AES models did not demonstrate strong gender biases, regardless of the LLM's accuracy in predicting students' genders. Additionally, the negative MAED values for both the "Unreliable" and "Correct" groups suggest that female students consistently experienced greater predictive errors than male students, irrespective of whether the LLM accurately identified their gender.

\begin{table*}[hbt!]
\begin{center}
\caption{AES accuracy and fairness results: The 'ns' label denotes non-significant results (p > 0.05). Higher QWK values indicate greater accuracy, while higher absolute values of bias metrics reflect greater bias. "Correct" signifies successful demographic prediction, whereas "Unreliable" indicates failed demographic prediction (including 'uncertain').}
\label{table3}
\resizebox{0.8\textwidth}{!}{
\begin{tabular}{@{}ccc|ccccc|ccccc@{}}
\toprule
\multirow{2}{*}{\begin{tabular}[c]{@{}c@{}}Essay\\ Set\end{tabular}} & \multirow{2}{*}{QWK} & \multirow{2}{*}{Correctness} & \multicolumn{5}{c|}{Language} & \multicolumn{5}{c}{Gender} \\ \cmidrule(l){4-13} 
 &  &  & QWK & OSA & OSD & CSD & MAED & QWK & OSA & OSD & CSD & MAED \\ \midrule
\multirow{2}{*}{1} & \multirow{2}{*}{0.456} & Unreliable & 0.352 & 0.168 & 0.150 & 0.082 & 0.654 & 0.455 & ns & ns & ns & -0.039 \\
 &  & Correct & 0.430 & 0.004 & 0.011 & 0.036 & -0.067 & 0.472 & ns & ns & ns & -0.020 \\ \midrule
\multirow{2}{*}{2} & \multirow{2}{*}{0.606} & Unreliable & 0.439 & 0.042 & ns & ns & -0.219 & 0.604 & ns & ns & ns & -0.056 \\
 &  & Correct & 0.598 & 0.023 & 0.015 & ns & 0.192 & 0.601 & ns & ns & ns & 0.165 \\ \midrule
\multirow{2}{*}{3} & \multirow{2}{*}{0.510} & Unreliable & 0.335 & ns & 0.040 & 0.046 & 0.171 & 0.507 & ns & ns & ns & -0.038 \\
 &  & Correct & 0.508 & ns & 0.066 & 0.095 & 0.042 & 0.485 & ns & ns & ns & 0.544 \\ \midrule
\multirow{2}{*}{4} & \multirow{2}{*}{0.522} & Unreliable & 0.240 & 0.045 & ns & ns & 0.243 & 0.517 & 0.004 & ns & ns & -0.076 \\
 &  & Correct & 0.508 & ns & ns & ns & -0.069 & 0.579 & ns & ns & ns & 0.048 \\ \midrule
\multirow{2}{*}{5} & \multirow{2}{*}{0.550} & Unreliable & 0.492 & ns & 0.080 & ns & 0.021 & 0.544 & 0.008 & ns & ns & -0.108 \\
 &  & Correct & 0.536 & ns & 0.021 & 0.023 & 0.051 & 0.543 & ns & ns & ns & 0.017 \\ \midrule
\multirow{2}{*}{6} & \multirow{2}{*}{0.532} & Unreliable & 0.257 & 0.041 & 0.068 & 0.051 & 0.253 & 0.543 & 0.004 & 0.009 & 0.009 & -0.109 \\
 &  & Correct & 0.561 & 0.009 & 0.021 & 0.093 & -0.170 & 0.367 & ns & ns & ns & -0.122 \\ \midrule
\multirow{2}{*}{Overall} & \multirow{2}{*}{0.629} & Unreliable & 0.485 & ns & ns & ns & 0.044 & 0.633 & 0.001 & ns & ns & -0.063 \\
 &  & Correct & 0.633 & 0.020 & 0.055 & 0.093 & -0.213 & 0.583 & ns & ns & ns & -0.055 \\ \bottomrule
\end{tabular}
}
\end{center}
\end{table*}

\begin{table*}[hbt!]
\begin{center}
\caption{Weighted multivariate regression results for scoring errors: The cell values represent coefficients, values in parentheses indicate standard errors, and starred values denote statistical significance (p < 0.05).}
\label{table4}
\resizebox{0.8\textwidth}{!}{
\begin{tabular}{@{}c|rrrr@{}}
\toprule
\multirow{2}{*}{\begin{tabular}[c]{@{}c@{}}Essay \\ Set\end{tabular}} & \multicolumn{4}{c}{\textbf{Gender}} \\ \cmidrule(l){2-5} 
 & Intercept & Gender & Correctness & Correctness*Gender \\ \midrule
1 & $^*0.591(0.056)$ & $-0.037(0.079)$ & $^*0.325(0.081)$ & $-0.052(0.111)$ \\
2 & $^*0.348(0.057)$ & $-0.101(0.080)$ & $-0.010(0.077)$ & $0.194(0.114)$ \\
3 & $^*0.141(0.055)$ & $-0.051(0.080)$ & $^*0.308(0.070)$ & $^*0.403(0.138)$ \\
4 & $-0.035(0.053)$ & $0.080(0.079)$ & $^*0.249(0.081)$ & $0.063(0.112)$ \\
5 & $^*-0.171(0.055)$ & $0.006(0.081)$ & $0.094(0.076)$ & $^*0.404(0.115)$ \\
6 & $^*0.831(0.043)$ & $-0.165(0.060)$ & $^*0.392(0.055)$ & $-0.058(0.092)$ \\
Overall & $^*0.343(0.023)$ & $-0.031(0.034)$ & $^*0.299(0.032)$ & $0.036(0.048)$ \\ \midrule
\multirow{2}{*}{\begin{tabular}[c]{@{}c@{}}Essay \\ Set\end{tabular}} & \multicolumn{4}{c}{\textbf{First-Language Background}} \\ \cmidrule(l){2-5} 
 & Intercept & Language & Correctness & Correctness*Language \\ \midrule
1 & $^*0.798(0.040)$ & $^*-0.548(0.196)$ & $^*-0.224(0.056)$ & $0.338(0.304)$ \\
2 & $^*0.346(0.043)$ & $^*0.301(0.115)$ & $-0.040(0.059)$ & $^*-0.490(0.203)$ \\
3 & $^*0.292(0.039)$ & $^*-0.292(0.113)$ & $^*-0.188(0.054)$ & $^*0.662(0.253)$ \\
4 & $0.051(0.042)$ & $-0.107(0.108)$ & $-0.024(0.058)$ & $0.080(0.262)$ \\
5 & $-0.015(0.039)$ & $^*-0.385(0.204)$ & $^*-0.175(0.055)$ & $^*0.612(0.277)$ \\
6 & $^*1.083(0.035)$ & $^*-0.479(0.070)$ & $^*-0.395(0.050)$ & $^*0.720(0.099)$ \\
Overall & $^*0.457(0.018)$ & $-0.084(0.048)$ & $^*-0.163(0.025)$ & $^*0.502(0.074)$ \\ \bottomrule
\end{tabular}
}
\end{center}
\end{table*}

Table \ref{table4} presents the weighted multivariate regression results for scoring errors. Gender was generally not a significant factor in scoring error variations, except in essay set 6, indicating that gender differences were not strongly associated with scoring errors in most cases. The Correctness variable was a significant predictor in essay set 1, 3, 4, and 6, with positive coefficients, indicating that when the LLM correctly predicted gender, it tended to make more scoring errors. Additionally, the interaction term Correctness*Gender was significant in essay set 3 and 5 with positive values, suggesting that scoring errors for male students increased when the LLM correctly identified them as male. In contrast, first-language background had a more substantial impact on scoring errors. The Language variable was significant in essay set 1, 2, 3, 5, and 6, indicating that variations in language background influenced scoring outcomes. Furthermore, the Correctness variable had a significant negative effect in Essay Sets 1, 3, 5, and 6, meaning that when the LLM correctly predicted a student's first-language background, it was associated with fewer scoring errors. The interaction term Correctness*Language was significant in Essay Sets 2, 3, 5, and 6, suggesting that the effect of the LLM correctly predicting first-language background varied depending on the first-language background. Notably, when analyzing the dataset as a whole, the positive coefficient of the Correctness*Language variable indicates that scoring errors for non-native English speakers increase when the LLM correctly identifies them as non-native.

\section{Discussion and Conclusions} \label{sec:disscuss}
To investigate the bias exhibited by LLMs in AES tasks and the relationship between such bias and LLMs' ability to predict students' demographics, this study evaluated GPT-4o's ability in inferring students' demographics. It analyzed how this predictive ability influenced the bias of the model’s scoring outcomes.

\noindent \textbf{Implications} Firstly, we demonstrated that prompt-based LLMs possess the capability to predict demographic attributes based on students' written essays. This finding aligns with prior research showing demographic attributes can be predicted from fine-tuned text embeddings of LLMs \cite{sha2022bigger,kwako2024can}. These findings contribute additional evidence that linguistic patterns are associated with demographic attributes, and AES models may inadvertently reinforce biases related to these attributes, potentially resulting in differential treatment of users based on implicit demographic cues. Secondly, our results indicate that scoring bias is more pronounced when the LLM accurately predicts students' first-language background. This finding aligns with previous research on fine-tuned LLMs, which has demonstrated that the predictive bias of downstream applications can stem from demographic information embedded in LLM representations. However, no such relationship was found for gender, which contradicts previous findings in fine-tuned LLMs \cite{sha2022bigger}. One potential reason for this discrepancy is the difference in methodology between fine-tuned and prompt-based LLMs. Fine-tuned models adjust their internal representations based on labeled training data, which can amplify demographic biases. In contrast, prompt-based LLMs rely on their pretrained knowledge and do not undergo task-specific fine-tuning, potentially reducing their sensitivity to certain demographic attributes. Additionally, first-language background may be more explicitly encoded in linguistic patterns within the model’s embeddings, whereas gender-related information might be less directly accessible in a prompt-based approach. Thirdly, debiasing strategies used in previous fine-tuning studies may not be directly applicable to prompt-based models. Earlier research typically relied on fine-tuning with carefully designed samples to reduce models' sensitivity to demographic attributes \cite{sha2022bigger,de2021stereotype}, thereby improving fairness in downstream tasks. However, this approach is impractical for prompt-based models if we rely solely on prompting, especially when it comes to the use of proprietary tools like ChatGPT. Nevertheless, it provides valuable insights such as the possibility of selecting few-shot examples with diverse demographic backgrounds to implicitly influence the model’s awareness of demographics and mitigate bias.

\noindent \textbf{Limitations} First, while we selected the state-of-the-art LLM model (GPT-4o), which has demonstrated superior performance in previous studies \cite{xiao2025human,stahl2024exploring}, our analysis was based on a single model. Results may vary with different LLMs, and we intend to extend our experiments to other advanced LLMs in future studies. Second, our study focused on gender and language background, but other demographic factors, such as race and socioeconomic status, warrant further investigation. Lastly, while we employed a well-established fairness evaluation framework, incorporating additional fairness metrics could offer a more comprehensive assessment of bias in AES.

%
%
%
\bibliographystyle{splncs04}
\bibliography{mybibliography}
\end{document}